%% file: main.tex
\documentclass[sigconf]{acmart}

\AtBeginDocument{%
  \providecommand\BibTeX{{%
    \normalfont B\kern-0.5em{\scshape i\kern-0.25em b}\kern-0.8em\TeX}}}

\acmYear{2023}
\setcopyright{acmlicensed}\acmConference[MM '23]{Proceedings of the 31st ACM International Conference on Multimedia}{October 29-November 3, 2023}{Ottawa, ON, Canada}
\renewcommand\footnotetextcopyrightpermission[1]{}
\usepackage{graphicx}
\usepackage{amsmath, bm}
\usepackage{amsthm}
\usepackage{booktabs}
\usepackage{algorithm}
\usepackage{algorithmic}
\usepackage{textcomp}
\usepackage{xcolor}
\usepackage{rotating}
\usepackage{multirow}
\usepackage{booktabs}
\usepackage{colortbl}
\usepackage{array}
\usepackage{overpic}
\usepackage{float}
\usepackage{balance}

\begin{document}

\title{M$^3$Net: Multi-view Encoding, Matching, and Fusion for Few-shot Fine-grained Action Recognition }

\author{Hao Tang}
\affiliation{%
  \institution{Nanjing University of Science and Technology}
  \state{Nanjing}
  \country{China}
}
\email{tanghao0918@njust.edu.cn}

\author{Jun Liu}
\affiliation{%
  \institution{Singapore University of Technology and Design}
  \country{Singapore}}
\email{jun_liu@sutd.edu.sg}

\author{Shuanglin Yan}
\affiliation{%
  \institution{Nanjing University of Science and Technology}
  \state{Nanjing}
  \country{China}
}
\email{shuanglinyan@njust.edu.cn}

\author{Rui Yan}
\affiliation{%
  \institution{Nanjing University}
  \state{Nanjing}
  \country{China}
}
\email{ruiyan@njust.edu.cn}

\author{Zechao Li}
\affiliation{%
  \institution{Nanjing University of Science and Technology}
  \state{Nanjing}
  \country{China}}
\email{zechao.li@njust.edu.cn}

\author{Jinhui Tang}\authornote{Corresponding author.}
\affiliation{%
  \institution{Nanjing University of Science and Technology}
  \state{Nanjing}
  \country{China}}
\email{jinhuitang@njust.edu.cn}

\renewcommand{\shortauthors}{Hao Tang et al.}

\begin{abstract}
Due to the scarcity of manually annotated data required for fine-grained video understanding, few-shot fine-grained (FS-FG) action recognition has gained significant attention, with the aim of classifying novel fine-grained action categories with only a few labeled instances. Despite the progress made in FS coarse-grained action recognition, current approaches encounter two challenges when dealing with the fine-grained action categories: the inability to capture subtle action details and the insufficiency of learning from limited data that exhibit high intra-class variance and inter-class similarity. To address these limitations, we propose M$^3$Net, a matching-based framework for FS-FG action recognition, which incorporates \textit{multi-view encoding}, \textit{multi-view matching}, and \textit{multi-view fusion} to facilitate embedding encoding, similarity matching, and decision making across multiple viewpoints. \textit{Multi-view encoding} captures rich contextual details from the intra-frame, intra-video, and intra-episode perspectives, generating customized higher-order embeddings for fine-grained data. \textit{Multi-view matching} integrates various matching functions enabling flexible relation modeling within limited samples to handle multi-scale spatio-temporal variations by leveraging the instance-specific, category-specific, and task-specific perspectives. \textit{Multi-view fusion} consists of matching-predictions fusion and matching-losses fusion over the above views, where the former promotes mutual complementarity and the latter enhances embedding generalizability by employing multi-task collaborative learning. Explainable visualizations and experimental results on three challenging benchmarks demonstrate the superiority of M$^3$Net in capturing fine-grained action details and achieving state-of-the-art performance for FS-FG action recognition.
\end{abstract}




\maketitle

\input{Introduction}
\input{RelatedWork}
\input{Method}
\input{Experiment}

% \begin{acks}
% This work was supported in part by the National Key Research and Development Program of China under Grant 2022ZD0118802, the National Natural Science Foundation of China under Grants 61932020 and U20B2064, the China Postdoctoral Science Foundation under Grant 2023TQ0151, and the Singapore Ministry of Education (MOE) AcRF Tier 2 under Grant MOE-T2EP20222-0009.
% \end{acks}

\bibliographystyle{ACM-Reference-Format}
% \balance
\bibliography{sample-base}

\end{document}

%% file: Introduction.tex
\section{Introduction}

Action recognition has achieved remarkable progress in video understanding, which is significantly driven by the introduction of large-scale datasets~\cite{russakovsky2015imagenet, SSV217, Kinetic17} and video models~\cite{TSN16, SlowFast19, MorphMLP22}.  However, the current approaches heavily depend on sufficient manually annotated samples~\cite{SunKRBWL23, 0018LL023}, which require significant labor and time resources to obtain, especially for fine-grained action categories. This bottleneck hinders further progress in video understanding. Thus, few-shot (FS) action recognition~\cite{CMN18, FuWFWBXJ19, AMeFuNet20} has attracted significant attention in addressing these constraints by reducing the need for manual annotations. The objective is to classify an unseen query video by assigning it to one of the action categories within the support set, even with limited annotated samples per action category. 
In this paper, we investigate the few-shot action recognition task in a more challenging setting, \emph{i.e.,~} few-shot fine-grained (FS-FG) action recognition,
where only a few or even one labeled sample is available to recognize novel fine-grained actions.

\begin{figure}[t!]
\centering
\includegraphics[width=\linewidth]{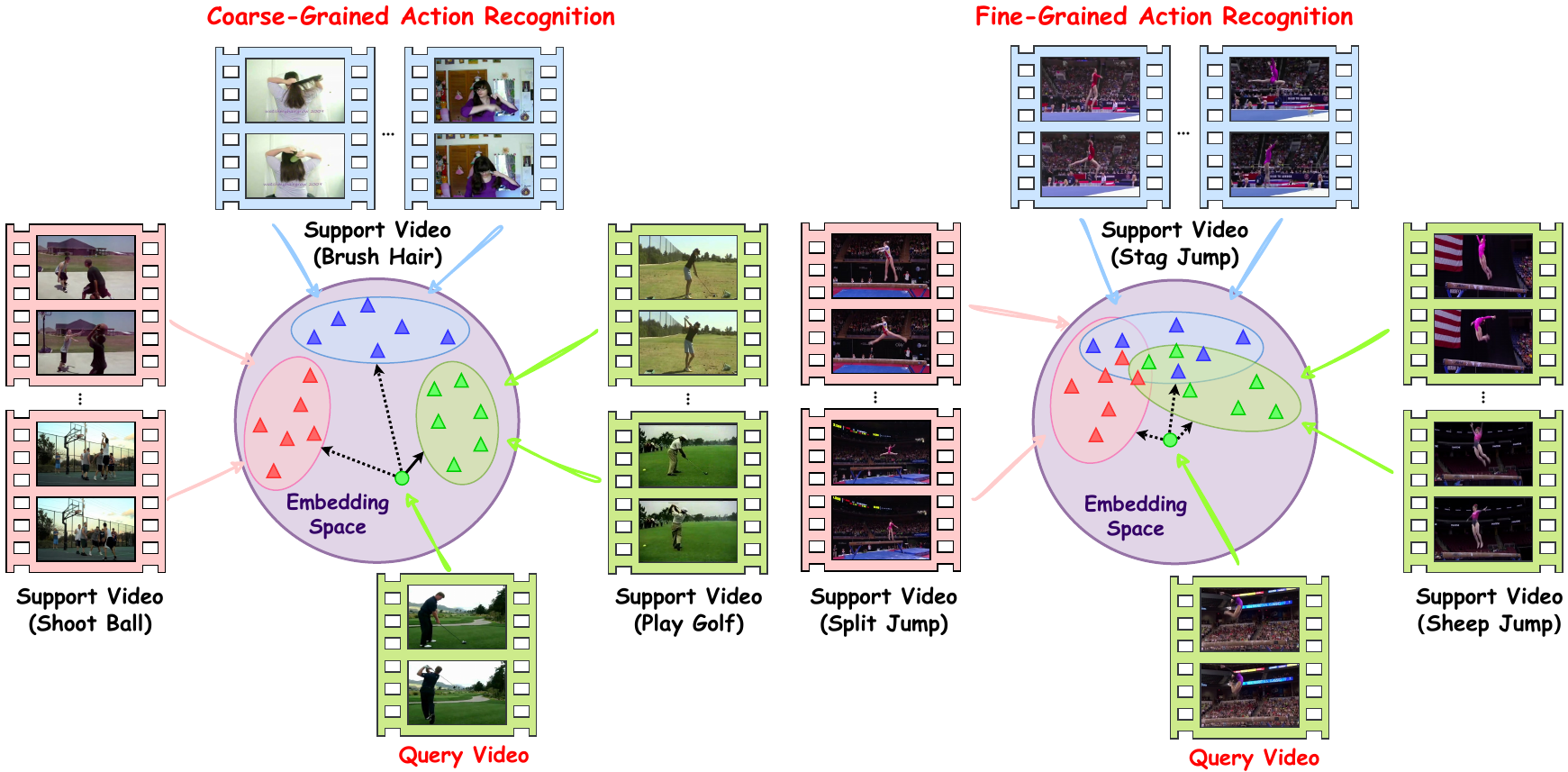}
% \vspace{-6mm}
\caption{\small Coarse vs. fine-grained action recognition in the 3-way 2-shot setting. \textbf{Left:} 
Coarse-grained action recognition requires differentiating visual appearance cues of objects and backgrounds extracted from limited frames.
\textbf{Right:} 
Fine-grained action recognition requires robust temporal reasoning at varying temporal scales and effective attention to fine details within limited frames 
due to subtle inter-class differences in the pose, specific sequence, and duration.
}%
\label{fig1:first_image}
% \vspace{-2mm}
\end{figure}

Recent efforts have yielded great success in FS image classification~\cite{PengLZLQT19, tang2020blockmix, TangYLT22, BSFA2022}, thereby prompting attempts to extend this paradigm to the domain of action recognition for FS video classification~\cite{Zhu00W21, LiLQLSFYL22, HuangYS22, ZhengCJ22}. 
These methods focus on coarse-grained action categories with different visual appearances, \emph{e.g.,~} ``Shoot Ball'' vs. ``Brush Hair'' in Fig.~\ref{fig1:first_image}.
Discriminating between these categories can be achieved via only background context, which can play a crucial role, sometimes even more significant than the action itself~\cite{LinLMLLXL022, LiXLY22, LiLLL23}.
Nevertheless, the need for action recognition at finer granularities is increasingly apparent in real-world scenarios such as sports analytics~\cite{YanXTS020, YanXSZT23, YanXTST23} and video surveillance~\cite{WangPWW22, ShenZ0XH22, ShenDZT23, ShenXZZZ23}, which usually requires a detailed comparison between similar actions with only subtle inter-class differences, \emph{e.g.,~} ``Split Jump'' vs. ``Sheep Jump'' in Fig.~\ref{fig1:first_image}.
Notably, the presented fine-grained action examples from FineGym~\cite{FineGym20} involving rapid body movements and drastic deformation, where the subtle differences only occur in the poses of certain sub-action rather than the visual appearances of surrounding objects and backgrounds relied upon by coarse-grained action recognition.
Hence, compared to coarse-grained recognition, recognizing fine-grained actions is significantly more challenging because of the subtle and discriminative multiple-dimension variations that existing coarse-grained methods typically fail to detect. Critically, applying existing FS action recognition methods directly to solve the FS-FG action recognition problem is not suitable.

\begin{figure*}[t!]
\centering
\includegraphics[width=\linewidth]{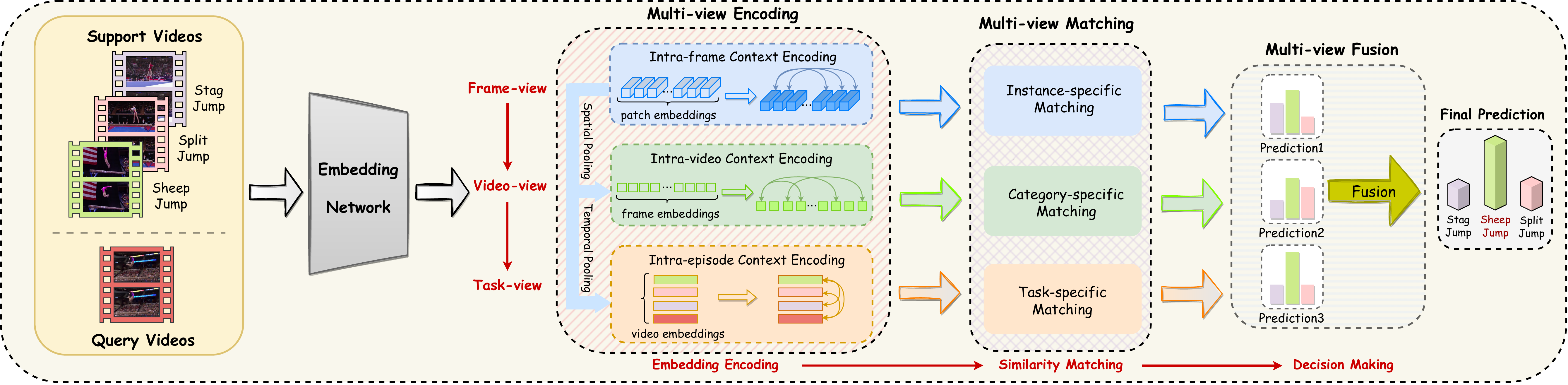}
% \vspace{-6mm}
\caption{\small Schematic illustration of the proposed M$^3$Net. Initially, given an episode of fine-grained videos, an embedding network is employed to extract their feature vectors. Subsequently, the \textit{multi-view encoding} is utilized to capture contextual details and generate customized higher-order embeddings. The resulting embeddings are then employed in the \textit{multi-view matching} to provide flexible relation modeling and robust video matching. Finally, the matching predictions from multiple views are merged in the \textit{multi-view fusion} to make a robust decision.
}%
\label{fig2:overview}
% \vspace{-2mm}
\end{figure*}

To address the challenges presented by both \textit{few-shot learning} and \textit{fine-grained recognition} in the FS-FG action recognition, we delve into this specific task from three distinct aspects: 
1) \textit{\bf Discriminative Embedding.} Given the large intra-class variance and subtle inter-class difference, capturing subtle spatial semantics and complicated temporal dynamics while minimizing unnecessary visual cues~\cite{MANet22} is critical for successful FS-FG action recognition. We argue that this requires encoding rich contextual details to generate different customized embeddings from multiple views for a fine-grained action. 
2) \textit{\bf Robust Matching.} Fine-grained actions involving various subactions with different speeds, orders, and offsets present difficulties in characterizing intricate spatio-temporal relations between a query video and limited support actions. Ideally, a more robust matching function should be employed from multiple dimensions to address the misalignment problem that arises from multi-scale spatio-temporal variations, as single-view temporal alignment metrics in existing methods may not suffice.
3) \textit{\bf Good Generalization.} To achieve optimal performance in few-shot learning, desirable models must transfer visual knowledge from the seen categories to the unseen categories, thereby tailoring task-specific discriminative representations~\cite{ETNDNet} for target actions, particularly those with high intra-class variance and inter-class similarity. As such, good generalization is a critical concern for the FS-FG action recognition task.

Motivated by the aforementioned observations, we propose M$^3$Net, a matching-based framework for FS-FG action recognition. M$^3$Net consists of \textbf{multi-view encoding}, \textbf{multi-view matching}, and \textbf{multi-view fusion}, which jointly facilitates embedding learning, similarity matching, and decision making across multiple views.
In the \textit{multi-view encoding}, we argue that the relevant relationships within a frame, a video, and an episode are desirable to generate customized features that are discriminative for intra-class and inter-class variance.
Initially, an \textit{intra-frame context encoding} (IFCE) module captures  subtle spatial semantics within a frame, allowing for the learning of instance-specific embeddings for fine-grained actions.
Subsequently, an \textit{intra-video context encoding} (IVCE) module captures complex temporal dynamics within a video, enabling learning of category-specific embeddings for fine-grained actions.
Meanwhile, an \textit{intra-episode context encoding} (IECE) module infers discriminative interactive clues within an episode, facilitating learning of task-specific embeddings for few-shot tasks.
The enriched higher-order embeddings from multiple views are then utilized in the \textit{multi-view matching}, which enables flexible relation modeling and robust video matching in the limited samples.
This is achieved using three proposed spatio-temporal matching functions: \textit{instance-specific}, \textit{category-specific}, and \textit{task-specific} matching.
Finally, we decouple \textit{multi-view fusion} of M$^3$Net into \textit{matching-losses fusion} and \textit{matching-predictions fusion} to guide the model to learn generalized embeddings and make a robust decision via a multi-task collaborative learning paradigm.
We evaluate M$^3$Net on three challenging fine-grained action recognition benchmarks (\emph{i.e.,}~Diving48~\cite{Diving48}, Gym99~\cite{FineGym20}, and Gym288~\cite{FineGym20}) and achieve remarkable performance improvements over current state-of-the-art methods.

Our contributions can be summarized as follows: 
(1) We propose a matching-based few-shot learning framework called M$^3$Net for fine-grained action recognition, which employs a multi-task collaborative learning paradigm combining \textit{multi-view encoding}, \textit{multi-view matching}, and \textit{multi-view fusion}.
(2) To generate customized representations with discriminative spatio-temporal cues, we propose a \textit{multi-view encoding} procedure that captures rich contextual details from the view of intra-frame, intra-video, and intra-episode.
(3) To address multi-scale spatio-temporal variations in fine-grained videos, we introduce three novel matching functions that model higher-order relations among limited samples in a \textit{multi-view matching} process.
(4) We instantiate the proposed \textit{multi-view fusion} in M$^3$Net as matching-losses fusion and matching-predictions fusion to foster cooperation and complementarity among multiple views.

%% file: RelatedWork.tex
\section{Related Work}
\subsection{Few-shot Learning}
Inspired by the data-efficient cognitive abilities of humans, the {few-shot learning} (FSL) task aims to learn new concept representations with limited supervised information~\cite{Wang21a}.
Recent research has investigated the FSL problem in various domains, including image classification~\cite{PengLZLQT19, KSTN2023}, object detection~\cite{AntonelliACCFGMMP22, KaulXZ22}, and segmentation~\cite{ZhangZT0S20, sun2022singular, ShiWLWWL22}. 
Existing FSL methods commonly follow a promising meta-learning paradigm, seeking to extract task-level knowledge across different episodes~\cite{finn2017model} and generalize the learned meta-knowledge to previously unseen tasks. 
General FSL methods can be roughly divided into three main categories based on the type of meta-knowledge acquired: optimization-based, metric-based, and augmentation-based methods.  
Optimization-based approaches~\cite{finn2017model, sun2019meta, jamal2019task} aim to learn optimal initialization parameters that enable quick adaptation to novel tasks with limited update steps.
Augmentation-based approaches~\cite{hariharan2017low, tang2020blockmix, schwartz2018delta} seek to increase the number of training samples or improve the diversity of feature distributions to augment model training.
Metric-based approaches~\cite{vinyals2016matching, snell2017prototypical, sung2018learning, TianTD21} aim to learn an embedding space and compare the similarity between query and support images using different distance metrics.
Our work is inspired by the metric-based ProtoNet~\cite{snell2017prototypical}, which builds robust category prototypes for each category and makes label predictions using nearest neighbor search. We share the same insight in a more challenging FS-FG action recognition task but focus on spatial-to-temporal relations modeling and matching across multiple views.

\subsection{Few-shot Action Recognition} 
Few-shot (FS) action recognition aims at learning to classify an unseen query action into one of the support action categories with extremely limited annotated samples~\cite{QianWYHW22, QianWHW23, qian2023adaptive}, which has attracted much attention recently. However, it differs from the previous FSL approaches, which are designed for two-dimensional images. This is because the task of FS action recognition takes place in a higher-dimensional space that incorporates temporal information from video frames.
The earliest research of FS action recognition could be traced back to CMN~\cite{CMN18}, which proposed a compound memory network employing a multi-saliency embedding algorithm to improve video representations for matching. Today, research in this field primarily concentrates on the metric-based meta-learning paradigm that can be categorized into two main directions: \textit{aggregation-based} and \textit{matching-based} methods. The former group~\cite{CMN18, TARN19, ARN20, CaoLLWZ21} concerns itself with the semantic modeling of context to produce video-level representations that are used for determining similarities. The latter group~\cite{OTAM20, ITANet21, TRX21, HRGSM22} focuses on explicit or implicit temporal context modeling for aligning frame-level sequences for the ultimate video matching. However, these methods concentrate solely on frame-wise matching, resulting in a lack of focus on higher-level temporal relations among multiple frames.
Thus, some methods~\cite{TRX21, STRM22} propose to align tuples of different sub-sequences utilizing enriched higher-order temporal representations to establish query-specific video matching. In addition to standard RGB frames, some works also utilize additional input to explicitly capture temporal clues, such as depth data~\cite{AMeFuNet20} and compressed domain data~\cite{LuoLLHML22}. Notably, unlike previous FS action recognition methods that focus primarily on {coarse-grained} action categories, this study is concerned with \textit{fine-grained} action categories. As a result, it is less optimal to apply them directly to FS-FG action recognition, as the scarcity of fine-grained data may result in overfitting. In this paper, we propose a departure from the established paradigm by exploiting the \textit{structural invariance} of multiple views for fine-grained actions. 
This allows FS-FG action recognition to be approached as a multi-task collaborative learning task with spatial-to-temporal context modeling, receiving added advantages from the mutual complementary of multi-view matching functions.

%% file: Method.tex
\section{Method}
\begin{figure*}[t!]
\centering
\includegraphics[width=\linewidth]{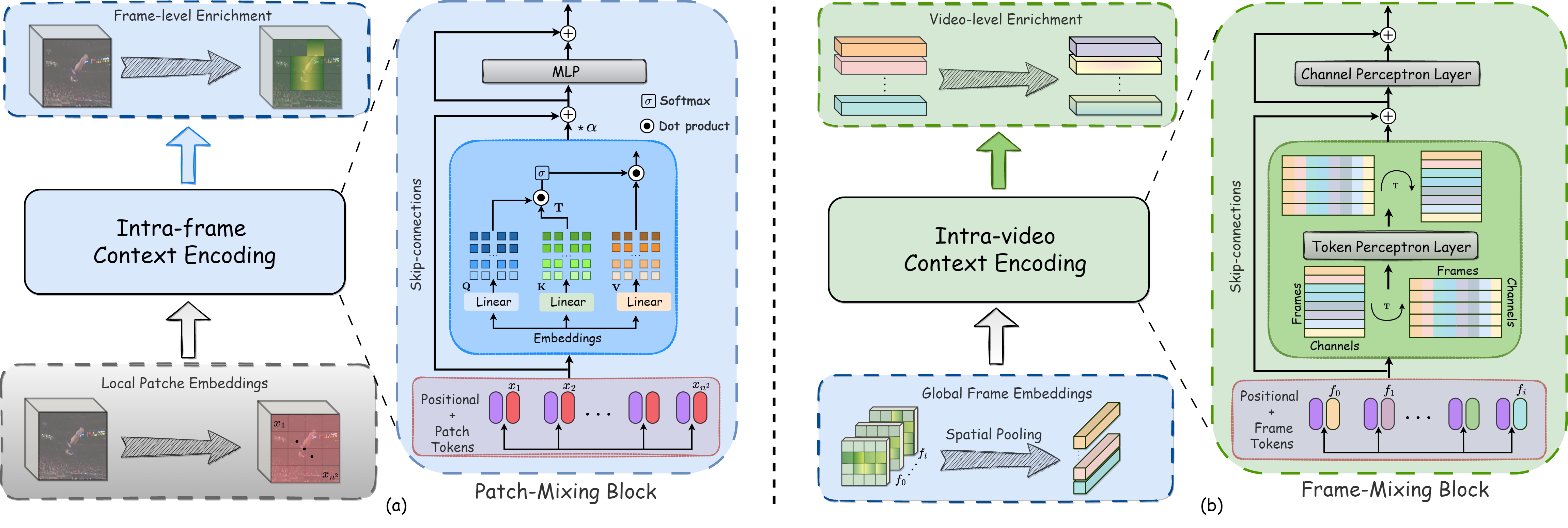}
% \vspace{-6mm}
\caption{\small Illustration of the proposed Intra-frame Context Encoding module and Intra-video Context Encoding module.}%
\label{fig3:spatial}
% \vspace{-2mm}
\end{figure*}
\subsection{Problem Formulation}
Following the conventional FS image recognition setup~\cite{tang2020blockmix,TangYLT22,KSTN2023}, the fine-grained
action recognition dataset is separated into a base set ${D}_\mathrm{base}=\{(x_{i}, y_{i})|y_{i}\in C_\mathrm{base}\}$ and a novel set $D_\mathrm{novel}=\{(x_{i}, y_{i})|y_{i}\in C_\mathrm{novel}\}$, wherein $y_{i}$ represents the action label of video $x_{i}$ and the base and novel action categories are disjoint, \emph{i.e.,~}$C_\mathrm{base}\cap C_\mathrm{novel}=\varnothing$. 
The goal of FS-FG action recognition involves training a model or optimizer on $D_\mathrm{base}$, which contains abundant training samples per base category, and subsequently, generalizing its performance to classify an \emph{unseen} query action in $D_\mathrm{novel}$ using only a few annotated support samples/shots.
Formally, in the $N$-way $K$-shot setting, a few-shot task (\emph{i.e.,~episode}) sampled from $D_\mathrm{novel}$ comprises of a support set $\mathcal{S}$ and a query set $\mathcal{Q}$. The support set consists of $N$ different action categories, \emph{i.e.,~}$C_\mathrm{support}\subseteq C_\mathrm{novel}$. Each category comprises $K$ support video clips, where $K$ typically ranges from $1$ to $5$. 
Thus, the objective is to recognize a query video $\bm{V}_{j}\in\mathcal{Q}$ as one of the support action categories with the aid of support samples $\mathcal{S}=\{\bm{V}_{1},\cdots, \bm{V}_{N\times K}\}$.
Consistent with previous methods~\cite{OTAM20, STRM22, HRGSM22}, we adopt an episodic training paradigm~\cite{vinyals2016matching} to sample numerous training episodes, which guarantees faithful training in the test environment.
% \vspace{-2mm}
\subsection{Overall Framework}
As illustrated in Fig.~\ref{fig2:overview}, we propose M$^3$Net, a matching-based few-shot learning framework that enables fine-grained recognition through a multi-task collaborative learning paradigm. Our design is to achieve complementary embedding learning, similarity matching, and metric-decision making in a three-step process: \textit{multi-view encoding}, \textit{multi-view matching}, and \textit{multi-view fusion}.

\noindent
{\bf Multi-view encoding:} To address intra-class and inter-class variance in fine-grained actions, we introduce the \textit{intra-frame context encoding} (IFCE) module as a spatial pathway to supplement instance-specific semantic information using patch-level spatial context within a frame. Additionally, we introduce the \textit{intra-video context encoding} (IVCE) module as a temporal pathway to dynamically capture motion-specific temporal context within a video. Our approach also includes the \textit{intra-episode context encoding} (IECE) module, serving as a task-adaptive pathway to capture discriminative task-specific cues across videos in an episode, mitigating inter-class variance of fine-grained actions.

\noindent
{\bf Multi-view matching:} 
Three novel spatio-temporal matching functions are suggested in M$^3$Net, namely \textit{instance-specific}, \textit{category-specific}, and \textit{task-specific} matching, based on video embeddings enhanced by the IFCE, IVCE, and IECE modules. To handle multi-scale spatio-temporal variations in fine-grained actions, we first propose the \textit{instance-specific} matching function for frame-level enriched features. For video-level enriched features and task-level enriched features, the proposed \textit{category-specific} and \textit{task-specific} matching functions make full use of limited available samples to facilitate flexible and robust video matching.

\noindent
{\bf Multi-view fusion:} To encourage generalized embeddings and improve decision-making abilities in a multi-task learning framework, we integrate diverse losses and predictions of multiple matching branches presented in the previous {multi-view matching} process. Empirically, the proposed \textit{multi-view fusion} procedure amplifies matching diversity and enhances the embedding generalization, ultimately benefitting FS-FG action recognition.

% \vspace{-2mm}
\subsection{Multi-view Encoding}
\subsubsection{\textbf{Intra-frame Context Encoding}}
The IFCE module aims to emphasize instance-specific semantics by utilizing spatial-based patch interaction within each frame of a fine-grained video. In essence, the IFCE module effectively manages the impact of class-irrelevant visual cues during feature transformation, consequently permitting subsequent \textit{instance-specific} matching.

Formally, let $\bm{f}_{i}\in \mathbb{R}^{h\times w\times d}$ denotes the feature map of $i$-\textit{th} frame in a video, comprising $h\times w$ patches with a $d$-dimensional embedding. 
Instead of using the entire spatial position~\cite{Wang0PT21}, an adaptive pooling operation is employed to transform $\bm{f}_{i}$ into $n\times n$ $(n<<h/w)$ non-overlapping local patches (also referred to as \textit{tokens}), 
% \emph{i.e.,~} $\left[f_{i}^{1}, f_{i}^{2}, \cdots, f_{i}^{n\times n} \right]$, 
which effectively limits the computation consumed by dot-product operation across the spatial dimension.
As shown in Fig.~\ref{fig3:spatial}(a), the core of the IFCE module is a \textit{patch-mixing block} that explicitly captures rich contextual information across all patches, with the sequence of patches $\bm{Z}_{i}=([\bm{f}_{i}^{1}, \bm{f}_{i}^{2}, \cdots, \bm{f}_{i}^{n\times n}]+\bm{P})\in \mathbb{R}^{n^{2}\times d}$ passed through it, where a learned positional embedding $\bm{P} \in \mathbb{R}^{{n^{2}}\times {d}}$ is added to the patches to retain positional information. 
To learn discriminative contextual information, the input $\bm{Z}_{i}$ is transformed  into ($\bm{Z}_i^q, \bm{Z}_i^k, \bm{Z}_i^v$) triplets using three matrices $\bm{W}_{Q}, \bm{W}_{K}, \bm{W}_{V}\in \mathbb{R}^{d\times d}$ as
$\bm{Z}_i^q=\bm{Z}_i\bm{W}_Q$, $\bm{Z}_i^k=\bm{Z}_i \bm{W}_K$, and $\bm{Z}_i^v=\bm{Z}_i \bm{W}_V$.
Given that $\bm{Z}_i^q$, $\bm{Z}_i^k$ and $\bm{Z}_i^v$ share the same input source, patch-level interaction can be described as:
\begin{equation}\label{eq2}
\hat{\bm{Z}}_i=\alpha \cdot \operatorname{Softmax}\left(\frac{\bm{Z}_i^q \cdot {\bm{Z}_i^k}^{\top}}{\sqrt{d}}\right) \cdot \bm{Z}_i^v + \bm{Z}_i,
\end{equation}
where $\operatorname{Softmax}(\cdot)$ is a row-wise softmax function used to determine the interaction weight between the patches, and $\alpha$ is an adaptive learning weight initialized to $1$ for residual learning. 
The pairwise similarity between patches in $\bm{Z}_i^q$ and $\bm{Z}_i^k$ defines the attention scores computed by the dot-product operation and is further used to produce the final weighted output $\bm{Z}_i^v$. 
The patch-attended feature $\hat{\bm{Z}}_i$ is passed through an MLP block to generate the final spatial context-aware feature as
% \begin{equation}
$\bm{f}_{i}^{'}=\operatorname{MLP}\left(\hat{\bm{Z}}_i\right)+\hat{\bm{Z}}_i$
% \end{equation}
where the MLP comprises three linear projections separated by two ReLU non-linearity.

\subsubsection{\textbf{Intra-video Context Encoding}}
The previously discussed IFCE module aims to effectively discover  subtle spatial semantics in the individual video frames. 
However, it encounters challenges when considering complex fine-grained actions that involve different orders and offsets. Because the IFCE module relies solely on the {spatial} perspective rather than considering multi-scale temporal variations from a {temporal} perspective. 
To address this limitation, we introduce an additional \textit{intra-video context encoding} (IVCE) module that functions as a temporal pathway, which is capable of adaptively capturing long-range temporal dynamics within a video and augmenting all frame features with their temporal context.

Formally, let the input video sequence $\bm{v}_{j} \in \mathbb{R}^{t \times n^{2} \times d}$ be enriched by the IFCE module for the $j$-\textit{th} video containing $t$ frames in an episode. We apply an average pooling operation to squeeze the spatial dimension of $\bm{v}_{j}$ to generate the frame-level global representations, denoted as $\bm{V}_j\in \mathbb{R}^{t\times d}$, and introduce a learned positional embedding $\bm{P} \in \mathbb{R}^{t\times d}$ applied to $\bm{V}_j$ to encode temporal order information among frames.
As illustrated in Fig.~\ref{fig3:spatial}(b), the core of the IVCE module is a \textit{frame-mixing block}, and all tokens are concatenated as input of this block to update the frame-level feature.
Specifically, a \textit{token perception layer}, which first applies to a transposed input table $\bm{V}_j^{\top}$ for {frame-mixing refinement}, is shared across the token dimension $d$ and followed by a \textit{channel perception layer} for {channel-mixing refinement}, which is shared across the frame tokens $t$. 
Both layers consist of two fully-connected layers separated by a ReLU non-linearity.
The whole process can be written as follows:
\begin{equation}\label{eq2}
\begin{aligned}
& \hat{\bm{V}}_j=\operatorname{ReLU}\left(\bm{V}_j^{\top} \bm{W}_{t_1}\right) \bm{W}_{t_2}+\bm{V}_j^{\top}, \\
& \bm{V}_{j}^{'}=\operatorname{ReLU}\left(\hat{\bm{V}}_j^{\top} \bm{W}_{c_1}\right) \bm{W}_{c_2}+\hat{\bm{V}}_j^{\top},
\end{aligned}
\end{equation}
where $\bm{W}_{t_1}, \bm{W}_{t_2}\in \mathbb{R}^{t\times t}$ and $\bm{W}_{c_1}, \bm{W}_{c_2}\in \mathbb{R}^{d\times d}$ are four learnable weights.
The output enriched video sequence feature is denoted as $\bm{V}_{j}^{'}\in \mathbb{R}^{t\times d}$, which is obtained by reshaping the temporally-enriched features back to the original token's dimensions. The designed \textit{token perception layer} and \textit{channel perception layer} allow interaction of different tokens and communication between different channels to refine each frame with temporal relational context.
%

% \vspace{-1mm}
\begin{figure}[t!]
\centering
\includegraphics[width=\linewidth]{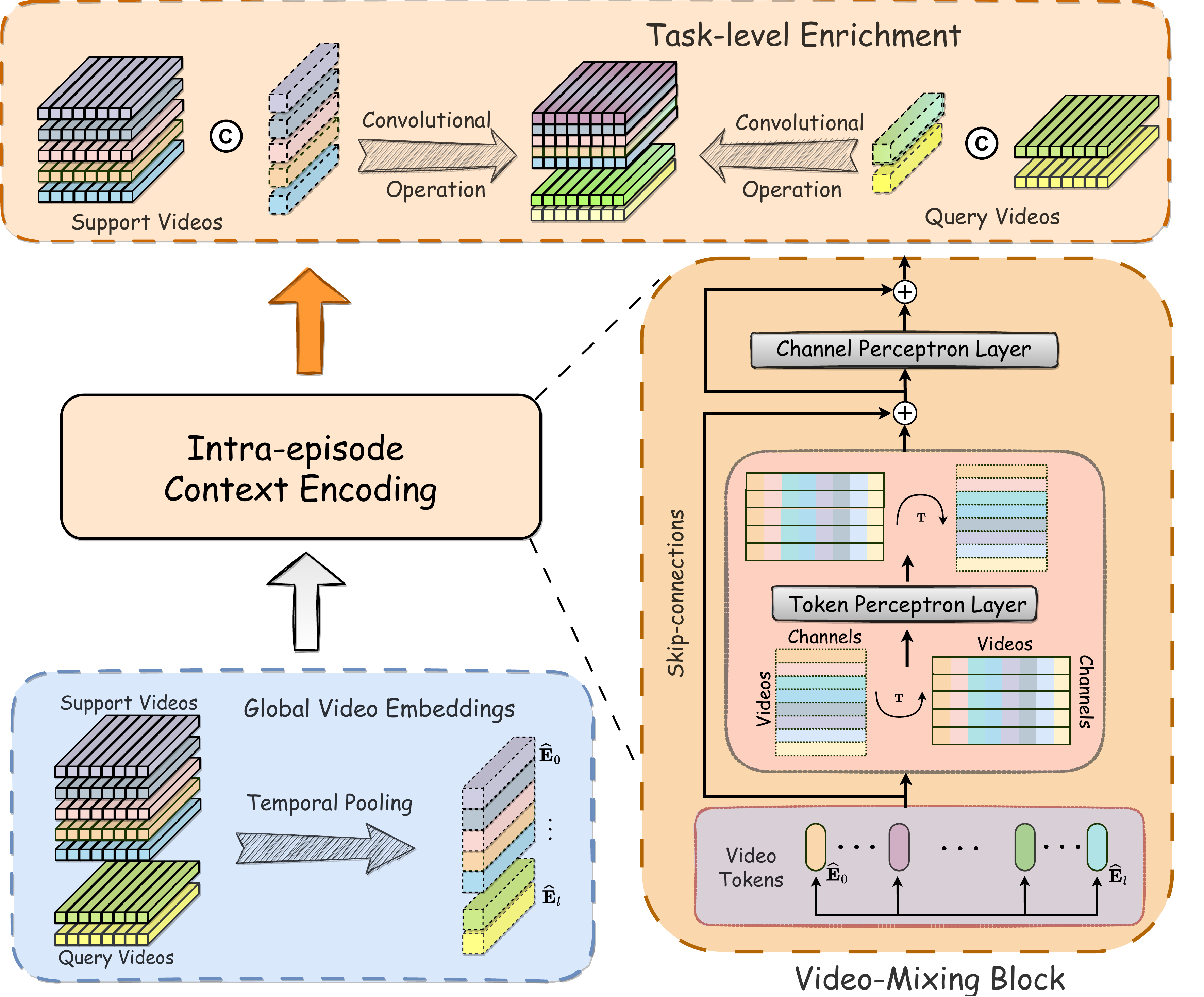}
% \vspace{-6mm}
\caption{\small Illustration of the proposed Intra-episode Context Encoding module.}%
\label{fig4:task}
% \vspace{-2mm}
\end{figure}

% \vspace{-1mm}
\subsubsection{\textbf{Intra-episode Context Encoding}}
The IFCE and IVCE modules enhance the original video representation in both spatial and temporal dimensions, highlighting class-specific representations and reducing {intra-class variance}. However, due to the small {inter-class variance} of fine-grained actions, the above {task-agnostic} embeddings are susceptible to overfitting the cues of the seen categories, which may impede good generalization in FS-FG action recognition. To mitigate this, we introduce the IECE module as a task-adaption pathway that captures discriminative interactive clues across videos. This allows for co-adaptation between videos in an episode to obtain {task-specific} embeddings.

Fig.~\ref{fig4:task} illustrates the core of the IECE module, namely \textit{video-mixing block}. This shares the same architecture as the \textit{frame-mixing block} in the IVCE module but differs in its usage and purpose.
Given an input episode (\emph{i.e.,~}a few-shot task) $\bm{e}$ containing the features of $l$ video clips, in which $t$ frames in each video are enriched ahead by the IFCE module, 
we first squeeze the spatial dimension of all features in $\bm{e}$ to obtain the global frame-level embeddings $\bm{E}\in \mathbb{R}^{l\times t\times d}$.
We then proceed to squeeze the temporal dimension of $\bm{E}$ to acquire the global video-level embeddings $\hat{\bm{E}}\in \mathbb{R}^{l\times d}$ for the entire episode.
Each video's global embedding $\hat{\bm{E}}_{i}\in \mathbb{R}^{d}$ is regarded as a token and we input all tokens to the \textit{video-mixing block} to update the video-level feature.
Since the \textit{token perception layer} in the \textit{video-mixing block} is insensitive to the order of the input videos, we cannot use position embedding for $\hat{\bm{E}}$.
Detailedly, the process of the \textit{video-mixing block} is similar to Eq.~\ref{eq2}, wherein the \textit{token perception layer} facilitates the interaction of all videos within an episode while considering the semantic relationships among different videos, and the \textit{channel perception layer} semantically contextualizes the video instances with relational context.
Finally, the output of the \textit{video-mixing block} serves as meta weights that help contextualize all video-level embeddings in $\bm{E}$, to enable strong co-adaptation of each item by a learnable \textit{convolution layer}. Therefore, the proposed IECE module achieves an implicit alignment effect on different videos, thus making all embeddings task-specific for subsequent task-specific matching.
% \vspace{-2mm}

\subsection{Multi-view Matching}\label{3.4}
The proposed \textit{multi-view encoding} in the preceding section captures intra-frame, intra-video, and intra-episode contexts to enhance the sub-sequence representations for matching. To ensure mutually complementary capabilities, customized matching functions are employed in \textit{multi-view matching} for the enriched video representations, which involves two types of matching functions, namely \textit{temporal} and \textit{non-temporal}, depending on the utilization of temporal information. 
Specifically, \textit{instance-specific} matching (I-M) as a \textit{temporal} function is utilized for embeddings enriched by the IFCE module, while \textit{category-specific} matching (C-M) and \textit{task-specific} matching (T-M) as two \textit{non-temporal} functions are adopted for embeddings enriched by the IVCE and IECE modules, respectively.
\vspace{-1mm}
\subsubsection{\textbf{Instance-specific Matching.}}
After enriching the embeddings with the IFCE module, we apply the proposed \textit{instance-specific} matching function to match the
query video $\bm{V}_j=\{\bm{f}_{y}^{j}\}_{y=1}^{t}\in \mathbb{R}^{t\times d}$ and support video $\bm{V}_i=\{\bm{f}_{x}^{i}\}_{x=1}^{t}\in \mathbb{R}^{t\times d}$,
where $t$ denotes the number of frames and $d$ indicates the dimension of frame-level embedding. 
Since the IFCE module does not incorporate temporal information, we align the frames of the two videos and infer the ordered temporal alignment score as a video-to-video similarity.
Based on inspired work from OTAM~\cite{OTAM20}, we first calculate the pairwise matching matrix $\bm{M}_{ij}\in \mathbb{R}^{t\times t}$, where $\bm{M}_{ij}(x,y)$ represents the cosine distance between the $x$-\textit{th} frame of video $\bm{V}_i$ and $y$-\textit{th} frame of video $\bm{V}_j$, \emph{i.e.,~}
$\bm{M}_{ij}(x,y) = 1-\frac{f_{x}^{i} \cdot f_{y}^{j}}
{\|f_{x}^{i}\| \|f_{y}^{j}\|}$.
Subsequently, we employ \textit{Dynamic Time Wrapping}~\cite{dynamic2007} for calculating the minimum cumulative matching costs over the units in $\bm{M}_{ij}$ via dynamic programming. Finally, we average the similarities over the optimal path with the minimum cumulative cost on $\bm{M}_{ij}$ to produce the video-to-video similarity, \emph{i.e.,~}$\bm{D}_{ij}\left(x, y\right)=\bm{M}_{ij}(x,y)+\min \left\{\bm{D}_{ij}\left(x-1, y-1\right), \bm{D}_{ij}\left(x-1, y\right), \bm{D}_{ij}\left(x, y-1\right)\right\}$,
where $1 \leq x \leq t$ and $1 \leq y \leq t$. We average the pairwise video-to-video similarities between the query and all support videos, which yields the final score $\mathcal{D}_{1}^{c}$ for class $c$. 
% \vspace{-1mm}
\subsubsection{\textbf{Category-specific Matching.}}
After acquiring embeddings enriched by the IVCE module for the support video and query video, we construct a \textit{query-centered prototype reconstruction} through a {cross-attention} process from the query samples to the support prototype. 
Specifically, we stack all frame embeddings from each support video to obtain the class prototype $\bm{V}^{c}$ for each action class $c$.
Given a query video clip $\bm{V}_{j}\in \mathbb{R}^{t\times d}$, the reconstructed prototype is formulated as 
$\hat{\bm{V}}^{c}=\operatorname{Softmax}\left(\frac{\bm{V}_{j} \bm{W}_{Q} (\bm{V}^{c}\bm{W}_{K})^{\top}}{\sqrt{d_k}}\right) \bm{V}^{c}\bm{W}_{V}$,
where $\bm{W}_{Q}/\bm{W}_{K}/\bm{W}_{V}\in \mathbb{R}^{d\times d_k}$.
Next, the distances between the frame embeddings of a query video and the reconstructed $c$-\emph{th} class prototype are aggregated to obtain the distance as $\mathbf{D}_c^{q\rightarrow s} = \left\|\bm{V}_{j}\bm{W}_{V}-\hat{\bm{V}}^{c}\right\|$.
However, focusing solely on the {query-specific prototype} is suboptimal for making full use of limited training data. Therefore, we propose a \textit{prototype-centered query reconstruction} by applying the same {cross-attention} process from the support set to the query set. 
Then, the distances between the reconstructed frame embeddings of the query video and the $c$-\emph{th} class prototype are aggregated to obtain $\mathbf{D}_c^{s\rightarrow q}$. Finally, the sum of the two distances is used for classifying the query action for class $c$, \emph{i.e.,~}$\mathcal{D}_2^{c}=\mathbf{D}_c^{s\rightarrow q}+\mathbf{D}_c^{q\rightarrow s}$.
% \vspace{-1mm}
\subsubsection{\textbf{Task-specific Matching.}}
Given the support video $\bm{V}_i$ and the query video $\bm{V}_j$, let $\bm{V}_i=\{\bm{f}_{x}^{i}\}_{x=1}^{t}, \bm{V}_j=\{\bm{f}_{y}^{j}\}_{y=1}^{t}\in \mathbb{R}^{t\times d}$ denote the sets of clip features enriched by the IECE module. Considering that both the support and query videos are contextualized in a task-specific manner, which achieves an implicit alignment effect, each frame in the query set is matched with its closest frame in the support set, and all query frame scores are averaged to obtain the final video-to-video similarity as $\mathbf{D}_{j}^{i}=\frac{1}{t} \sum_{\bm{f}_x^i \in \bm{V}_i}\left(\min _{\bm{f}_y^j \in \bm{V}_j}\left\|\bm{f}_x^i-\bm{f}_y^j\right\|\right)$. Similar to \textit{category-specific} matching function, we derive the symmetric process in this function where every frame from the support set must match jointly as $\mathbf{D}_{i}^{j}=\frac{1}{t} \sum_{\bm{f}_y^j \in \bm{V}_j}\left(\min _{\bm{f}_x^i \in \bm{V}_i}\left\|\bm{f}_y^j-\bm{f}_x^i\right\|\right)$. The similarity score between the two videos is calculated as $\mathbf{D}_{ij} = \mathbf{D}_{j}^{i}+\mathbf{D}_{i}^{j}$.
We average the similarity scores between the query video and all support videos in class $c$ to obtain the final score $\mathcal{D}_{3}^{c}$ for this class. 
% \vspace{-1mm}
\subsection{Multi-task Learning with Multi-view Fusion}
To strengthen the collaborative power of \textit{multi-view encoding} and amalgamate the contributions of diverse \textit{multi-view matching} functions, we reformulate the optimization of the proposed framework as a multi-task collaborative learning paradigm and boost performance from a \textit{multi-view fusion} perspective, as displayed in Fig.~\ref{fig2:overview}.

In doing so, we decouple the proposed \textit{multi-view fusion} of M$^3$Net into \textit{matching-losses fusion} and \textit{matching-predictions fusion} to facilitate the model in learning generalized embeddings and to make a robust decision to accomplish complementarity among multiple views.
During training, given the ground-truth class labels $\mathbf{Y}^{q} \in \mathbb{R}^{N}$ for $N$-way $K$-shot settings, the similarity scores $\mathcal{D}_{1}, \mathcal{D}_{2}, \mathcal{D}_{3}$ from Sec.~\ref{3.4} are passed through \emph{softmax} to obtain class probabilities $\mathbf{Y}_{1}, \mathbf{Y}_{2}, \mathbf{Y}_{3} \in \mathbb{R}^{N}$ , which are individually optimized via cross-entropy loss.
Thus, the classification loss for each matching branch can be formulated as 
$\mathcal{L}_1 = \operatorname{CE}(\mathbf{Y}_{1}, \mathbf{Y}^{q})$, $\mathcal{L}_2 = \operatorname{CE}(\mathbf{Y}_{2}, \mathbf{Y}^{q})$, and $\mathcal{L}_3 = \operatorname{CE}(\mathbf{Y}_{3}, \mathbf{Y}^{q})$,
where $\operatorname{CE}(\cdot)$ refers to the cross-entropy loss function. As a result, the comprehensive \textit{matching-losses fusion} is delineated as $\mathcal{L} = \mathcal{L}_1+\mathcal{L}_2+\mathcal{L}_3$ for learning generalized spatial-temporal embeddings. At the inference stage, we construct a comprehensive multi-view predictor by summing the prediction distribution of individual matching branches as follows:  $\mathbf{Y}=\mathbf{Y}_{1}+\mathbf{Y}_{2}+\mathbf{Y}_{3}.$

%% file: Experiment.tex
\section{Experiments}
\begin{table*}[th!] 
\centering
\caption{\small Comparison of 1-shot, 3-shot, and 5-shot recognition performance on the Diving48, Gym99 and Gym288 datasets.}
\label{tab:compare_SOTA}
\vspace{-1mm}
\resizebox{\linewidth}{!}{
\begin{tabular}{l|c|ccc|ccc|ccc}
\toprule
\multicolumn{1}{l}{} & & \multicolumn{3}{c|}{\bf Diving48}  & \multicolumn{3}{c|}{\bf Gym99} & \multicolumn{3}{c}{\bf Gym288}\\
\midrule
\hspace{-0mm} {\bf Method} \hspace{2mm} &  \hspace{1mm} {\bf Reference} \hspace{1mm} &  1-shot  &   3-shot & 5-shot &  1-shot  & 3-shot & 5-shot & 1-shot  & 3-shot & 5-shot \\
\midrule
\hspace{-0mm} {TSN\cite{TSN16}+Cosine}  & ECCV'16 &   {29.72}~{\footnotesize $\pm$ 0.25}  & {35.69}~{\footnotesize $\pm$ 0.26} & {38.81}~{\footnotesize $\pm$ 0.26} & {36.63}~{\footnotesize $\pm$ 0.30} & {41.35}~{\footnotesize $\pm$ 0.31} & {42.76}~{\footnotesize $\pm$ 0.31} & {36.68}~{\footnotesize $\pm$ 0.29} & {42.87}~{\footnotesize $\pm$ 0.29} & {44.82}~{\footnotesize $\pm$ 0.30}\\ 
\hspace{-0mm} {ProtoNet} \cite{snell2017prototypical}  & NeurIPS'17 &   {59.80}~{\footnotesize $\pm$ 0.33}  & {74.74}~{\footnotesize $\pm$ 0.29} & {77.69}~{\footnotesize $\pm$ 0.28} & {62.99}~{\footnotesize $\pm$ 0.36} & {74.25}~{\footnotesize $\pm$ 0.31} & {77.99}~{\footnotesize $\pm$ 0.30} & {58.98}~{\footnotesize $\pm$ 0.35} & {70.56}~{\footnotesize $\pm$ 0.31} & {74.87}~{\footnotesize $\pm$ 0.30}\\ 
\hspace{-0mm} OTAM \cite{OTAM20}   & CVPR'20 & {41.64}~{\footnotesize $\pm$ 0.29}  & {47.96}~{\footnotesize $\pm$ 0.27} & {49.78}~{\footnotesize $\pm$ 0.26} & {48.93}~{\footnotesize $\pm$ 0.35} &  {52.88}~{\footnotesize $\pm$ 0.32} & {55.61}~{\footnotesize $\pm$ 0.32}&{47.58}~{\footnotesize $\pm$ 0.33}&{53.67}~{\footnotesize $\pm$ 0.31}& {55.62}~{\footnotesize $\pm$ 0.31}\\ 
\hspace{-0mm} {PAL} \cite{PAL21}  & BMVC'21 &   {50.40}~{\footnotesize $\pm$ 0.24}  & {57.57}~{\footnotesize $\pm$ 0.27} & {61.17}~{\footnotesize $\pm$ 0.27} & {58.85}~{\footnotesize $\pm$ 0.31} & {65.13}~{\footnotesize $\pm$ 0.32} & {68.64}~{\footnotesize $\pm$ 0.33} 
& {58.32}~{\footnotesize $\pm$ 0.32} & {63.62}~{\footnotesize $\pm$ 0.34} & {65.77}~{\footnotesize $\pm$ 0.35} 
\\ 
\hspace{-0mm} TRX \cite{TRX21}  & CVPR'21 & {62.53}~{\footnotesize $\pm$ 0.32}  & {77.81}~{\footnotesize $\pm$ 0.29}  & {82.01}~{\footnotesize $\pm$ 0.26} & {66.55}~{\footnotesize $\pm$ 0.33} & {80.15}~{\footnotesize $\pm$ 0.30}  &  {83.86}~{\footnotesize $\pm$ 0.27} & {64.03}~{\footnotesize $\pm$ 0.35} & {76.73}~{\footnotesize $\pm$ 0.30}& {81.05}~{\footnotesize $\pm$ 0.27} \\ 
\hspace{-0mm} STRM \cite{STRM22}  & CVPR'22 &  {62.29}~{\footnotesize $\pm$ 0.34} &  {77.79}~{\footnotesize $\pm$ 0.29}& {81.03}~{\footnotesize $\pm$ 0.27}  & {68.09}~{\footnotesize $\pm$ 0.34} & {80.93}~{\footnotesize $\pm$ 0.29} & {84.33}~{\footnotesize $\pm$ 0.26} & {64.81}~{\footnotesize $\pm$ 0.36} & {76.98}~{\footnotesize $\pm$ 0.31} & {80.64}~{\footnotesize $\pm$ 0.28}\\ %
\hspace{-0mm} {HyRSM} \cite{HRGSM22} & CVPR'22 &   {61.04}~{\footnotesize $\pm$ 0.30}  & {76.04}~{\footnotesize $\pm$ 0.30} & {81.74}~{\footnotesize $\pm$ 0.28} & {67.09}~{\footnotesize $\pm$ 0.33} & {78.15}~{\footnotesize $\pm$ 0.32} & {82.92}~{\footnotesize $\pm$ 0.31} & {62.86}~{\footnotesize $\pm$ 0.35} & {73.12}~{\footnotesize $\pm$ 0.32} & {78.81}~{\footnotesize $\pm$ 0.31}\\ 
\midrule
\hspace{-0mm} M$^3$Net  & Ours & \bf {72.35}~{\footnotesize $\pm$ 0.33} & \bf {82.50}~{\footnotesize $\pm$ 0.26}& \bf {85.44}~{\footnotesize $\pm$ 0.24}  & \bf {72.70}~{\footnotesize $\pm$ 0.33} & \bf {83.03}~{\footnotesize $\pm$ 0.27} & \bf {86.71}~{\footnotesize $\pm$ 0.24} & \bf {68.66}~{\footnotesize $\pm$ 0.34} & \bf {79.73}~{\footnotesize $\pm$ 0.29} & \bf {83.05}~{\footnotesize $\pm$ 0.26}\\ 
\bottomrule
\end{tabular}}
% \vspace{-2mm}
\end{table*}

\subsection{Datasets}
We evaluate the effectiveness of our proposed M$^3$Net against other competing methods on two common fine-grained action recognition datasets, namely FineGym~\cite{FineGym20} and Diving48~\cite{Diving48}, using $5$-way $1/3/5$-shot FS recognition tasks for all models considered. To ensure the reliability of the results, we devise carefully-planned evaluation protocols for both datasets. Specifically, we determine the split of training, validation, and testing action categories, as well as the number of samples in each episode. 

\textbf{FineGym} provides annotations of hierarchical fine-grained actions in gymnastic events, which comprise two levels of fine-grained categories, namely \textbf{Gym99} and \textbf{Gym288}, with over 34k (99 classes) and 38k (288 classes) samples respectively. 
In the case of \textbf{Gym99}, a $61/12/26$ split is employed for the training/validation/testing categories.
Due to insufficient support samples for multi-shot tasks, some actions in \textbf{Gym288} are excluded, which results in the splits of training/validation/testing categories being set to $128/25/61$.

\textbf{Diving48} is a fine-grained video dataset of competitive diving, consisting of 18,404 trimmed video clips of 48 well-defined dive sequences. This proves to be a challenging task for fine-grained action recognition as dives may differ in different stages and thus require modeling of long-term temporal dynamics. Here, we randomly select $28$ training, $5$ validation, and $15$ testing categories.
% \vspace{-2mm}
\subsection{Implementation Details}
Following the previous works~\cite{OTAM20, STRM22, HRGSM22}, we employ ResNet-50~\cite{HeZRS16} as the embedding model and initialize it with weights pre-trained on the ImageNet dataset~\cite{russakovsky2015imagenet}. To accurately represent each fine-grained action, we uniformly and sparsely sample $8$ frames per video (\emph{i.e.,~}$t=8$) as in previous methods~\cite{STRM22, HRGSM22}. 
We employ an IFCE module by setting $n=4$, where $f_i$ has a dimension of $d=2048$.
During the training phase, the input video frames are randomly cropped to $224\times 224$, we introduce basic data augmentation techniques including random cropping and random horizontal flip.
The optimizer used to optimize our M$^3$Net is the SGD optimizer having an initialized learning rate of $1e-4$. The training procedure continues for $60,000$ episodes on all datasets with the learning rate decaying by $0.5$ after every $2,000$ episodes. During the inference phase, we resize all inputs to $256 \times 256$ before center cropping. We perform $5-$way $1/3/5-$shot evaluation for FS-FG action recognition on all datasets and report the mean accuracy obtained from $6,000$ episodes randomly selected from the testing set.
%
% \vspace{-2mm}

\begin{table}[t!]
\caption{\small Impact of the key components in multi-view encoding on Gym99 for 5-way 1-shot action recognition. To simplify, the intra-frame, intra-video, and intra-episode context encoding are represented as "I-F", "I-V", and "I-E", respectively. 
}\label{tab2}
\vspace{-1mm}
\resizebox{\linewidth}{!}{
\begin{tabular}{clllllllll}
\toprule
\multicolumn{2}{c|}{\bf Fusion Num} & \multicolumn{1}{c}{$1$} & \multicolumn{1}{c|}{$1$} & \multicolumn{1}{c}{$1$} & \multicolumn{1}{c}{$1$} & \multicolumn{1}{c|}{$1$} & \multicolumn{1}{c}{$1$} & \multicolumn{1}{c}{\bf $1$} & \multicolumn{1}{c}{\bf $1$}\\
\midrule
% IFCE
\multicolumn{1}{c|}{\multirow{2}*{\bf Multi-view}}&\multicolumn{1}{c|}{\bf I-F}& \multicolumn{1}{c}{-}&\multicolumn{1}{c|}{\bf \checkmark}&\multicolumn{1}{c}{-}& \multicolumn{1}{c}{-} & \multicolumn{1}{c|}{\bf \checkmark} &\multicolumn{1}{c}{-}& \multicolumn{1}{c}{-} & \multicolumn{1}{c}{\bf \checkmark}\\
% IVCE
\multicolumn{1}{c|}{\multirow{2}*{\bf Encoding}}&\multicolumn{1}{c|}{\bf I-V} & \multicolumn{1}{c}{-}&\multicolumn{1}{c|}{-}&\multicolumn{1}{c}{-}& \multicolumn{1}{c}{\bf \checkmark} & \multicolumn{1}{c|}{\bf \checkmark} &\multicolumn{1}{c}{-}& \multicolumn{1}{c}{-} & \multicolumn{1}{c}{-}\\
% ITCE
\multicolumn{1}{c|}{}&\multicolumn{1}{c|}{\bf I-E} & \multicolumn{1}{c}{-}&\multicolumn{1}{c|}{-}&\multicolumn{1}{c}{-}& \multicolumn{1}{c}{-} & \multicolumn{1}{c|}{-} &\multicolumn{1}{c}{-}& \multicolumn{1}{c}{\bf \checkmark} & \multicolumn{1}{c}{\bf \checkmark}\\
\midrule
% I-M
\multicolumn{1}{c|}{\multirow{2}*{\bf Multi-view}}&\multicolumn{1}{c|}{\bf I-M} &\multicolumn{1}{c}{\bf \checkmark}&\multicolumn{1}{c|}{\bf \checkmark}& \multicolumn{1}{c}{-} &\multicolumn{1}{c}{-}& \multicolumn{1}{c|}{-} & \multicolumn{1}{c}{-} & \multicolumn{1}{c}{-} & \multicolumn{1}{c}{-}\\
% C-M
\multicolumn{1}{c|}{\multirow{2}*{\bf Matching}}&\multicolumn{1}{c|}{\bf C-M} &\multicolumn{1}{c}{-}& \multicolumn{1}{c|}{-} &\multicolumn{1}{c}{\bf \checkmark}& \multicolumn{1}{c}{\bf \checkmark} &\multicolumn{1}{c|}{\bf \checkmark}& \multicolumn{1}{c}{-} & \multicolumn{1}{c}{-} & \multicolumn{1}{c}{-}\\
% T-M
\multicolumn{1}{c|}{}&\multicolumn{1}{c|}{\bf T-M} & \multicolumn{1}{c}{-}&\multicolumn{1}{c|}{-}&\multicolumn{1}{c}{-}& \multicolumn{1}{c}{-} & \multicolumn{1}{c|}{-} &\multicolumn{1}{c}{\bf \checkmark}& \multicolumn{1}{c}{\bf \checkmark} & \multicolumn{1}{c}{\bf \checkmark}\\
\midrule
\multicolumn{1}{c|}{\bf Fusion}&\multicolumn{1}{c|}{$\mathbf{Y}$} & \multicolumn{1}{c}{ {46.53}} & \multicolumn{1}{c|}{ \bf{50.09}} & \multicolumn{1}{c}{{64.66}} & \multicolumn{1}{c}{{68.92}} & \multicolumn{1}{c|}{ \bf{69.53}} & \multicolumn{1}{c}{{44.50}} & \multicolumn{1}{c}{{58.15}} & \multicolumn{1}{c}{\bf{59.79}}\\
\bottomrule
\end{tabular}}
\vspace{-2mm}
\end{table}

\subsection{Comparison with state-of-the-art}
This section presents a comparative evaluation of state-of-the-art FS action recognition methods for the standard $5-$way $1/3/5-$shot task on three fine-grained benchmarks, considering only methods employing a 2D embedding backbone (\emph{i.e,~}ResNet50) for per-frame features extraction.
Our proposed M$^3$Net significantly outperforms existing methods, setting a new state-of-the-art on all three benchmarks, as demonstrated in Tab.~\ref{tab:compare_SOTA}. 
On Diving48, TRX~\cite{TRX21}, STRM~\cite{STRM22}, and HyRSM~\cite{HRGSM22} achieve comparable recognition accuracies. In comparison, M$^3$Net outperforms existing methods with a higher performance of $72.35/82.50/85.44\%$ for $1/3/5-$shot. 
Notably, M$^3$Net achieves a significant performance improvement of up to $18.52\%$ in the most challenging 1-shot setting.
Moreover, the dataset Diving48 involves appearance-related scene understanding, while the Gym99 and Gym288 datasets primarily focus on motion-based temporal reasoning. 
M$^3$Net demonstrates outstanding performance on Gym99 and Gym288 under the $1/3/5-$shot settings, showcasing its ability to learn discriminative fine-grained differences from limited and similar samples.
Overall, M$^3$Net is more comprehensive in reflecting actual distances between videos and demonstrates strong robustness and generalization across these fine-grained datasets with superior performance on few-shot tasks. 

Observations from the results in Tab.~\ref{tab:compare_SOTA} indicate that: 
(1)
Coarse-grained action recognition models are unsuitable for fine-grained tasks, as M$^3$Net consistently outperforms the most advanced temporal alignment method~\cite{STRM22} and hybrid relation method~\cite{HRGSM22} for all shot settings.
(2) 
Although STRM~\cite{STRM22} utilizing enriched tuples outperforms other methods, it does not work well on FS-FG action recognition tasks as it utilizes sparsely sampled frame tuples as input without considering task-specific features, resulting in missing fine-grained action details. 
(3)
HyRSM~\cite{HRGSM22} shares similar insights with M$^3$Net to generate task-specific features. However, its performance is slightly behind TRX~\cite{TRX21} and STRM~\cite{STRM22}, as the learned task-specific feature in HyRSM is not discriminative for fine-grained actions and the single matching function is susceptible to noise interference.

\begin{table}[t!]
\caption{\small Comparison between different combinations of multi-view matching functions for 5-way 1/5-shot action recognition on Gym99. 
}\label{tab4}
\vspace{-1mm}
\scriptsize
\resizebox{\linewidth}{!}{
\begin{tabular}{cllllll}
\toprule
%L1
\multicolumn{1}{c|}{\multirow{2}*{\bf Intra-frame}}&\multicolumn{1}{c|}{\bf I-M}& \multicolumn{1}{c|}{\bf \checkmark}&\multicolumn{1}{c|}{-}&\multicolumn{1}{c|}{-}& \multicolumn{1}{c|}{\bf \checkmark} & \multicolumn{1}{c}{\bf \checkmark}\\
%L2
\multicolumn{1}{c|}{\multirow{2}*{\bf Context Encoding}}&\multicolumn{1}{c|}{\bf C-M} & \multicolumn{1}{c|}{-}&\multicolumn{1}{c|}{\bf \checkmark}&\multicolumn{1}{c|}{-}& \multicolumn{1}{c|}{-} & \multicolumn{1}{c}{-} \\
%L3
\multicolumn{1}{c|}{}&\multicolumn{1}{c|}{\bf T-M} & \multicolumn{1}{c|}{-}&\multicolumn{1}{c|}{-}&\multicolumn{1}{c|}{\bf \checkmark}& \multicolumn{1}{c|}{-} & \multicolumn{1}{c}{-} \\
\midrule
% Y1
\multicolumn{1}{c|}{\multirow{2}*{\bf Intra-video}}&\multicolumn{1}{c|}{\bf I-M} &\multicolumn{1}{c|}{\bf \checkmark}&\multicolumn{1}{c|}{-}& \multicolumn{1}{c|}{-} &\multicolumn{1}{c|}{-}& \multicolumn{1}{c}{-} \\
% Y2
\multicolumn{1}{c|}{\multirow{2}*{\bf Context Encoding}}&\multicolumn{1}{c|}{\bf C-M} &\multicolumn{1}{c|}{-}& \multicolumn{1}{c|}{ \bf \checkmark} &\multicolumn{1}{c|}{-}& \multicolumn{1}{c|}{} &\multicolumn{1}{c}{\bf \checkmark}\\
%Y3
\multicolumn{1}{c|}{}&\multicolumn{1}{c|}{\bf T-M} & \multicolumn{1}{c|}{-}&\multicolumn{1}{c|}{-}&\multicolumn{1}{c|}{ \bf \checkmark}& \multicolumn{1}{c|}{ \bf \checkmark} & \multicolumn{1}{c}{-} \\
\midrule
%
% Y1
\multicolumn{1}{c|}{\multirow{2}*{\bf Intra-episode}}&\multicolumn{1}{c|}{\bf I-M} &\multicolumn{1}{c|}{ \bf \checkmark}&\multicolumn{1}{c|}{-}& \multicolumn{1}{c|}{-} &\multicolumn{1}{c|}{-}& \multicolumn{1}{c}{-} \\
% Y2
\multicolumn{1}{c|}{\multirow{2}*{\bf Context Encoding}}&\multicolumn{1}{c|}{\bf C-M} &\multicolumn{1}{c|}{-}& \multicolumn{1}{c|}{ \bf \checkmark} &\multicolumn{1}{c|}{-}& \multicolumn{1}{c|}{ \bf \checkmark} &\multicolumn{1}{c}{-}\\
%Y3
\multicolumn{1}{c|}{}&\multicolumn{1}{c|}{\bf T-M} & \multicolumn{1}{c|}{-}&\multicolumn{1}{c|}{-}&\multicolumn{1}{c|}{ \bf \checkmark}& \multicolumn{1}{c|}{ -} & \multicolumn{1}{c}{\bf \checkmark} \\
\midrule
\multicolumn{1}{c|}{\bf Fusion (1-shot)}&\multicolumn{1}{c|}{$\mathbf{Y}$} & \multicolumn{1}{c|}{ {68.05}} & \multicolumn{1}{c|}{72.05} & \multicolumn{1}{c|}{68.13} & \multicolumn{1}{c|}{72.22} & \multicolumn{1}{c}{\bf{72.70}}\\
% \hline
\multicolumn{1}{c|}{\bf Fusion (5-shot)}&\multicolumn{1}{c|}{$\mathbf{Y}$} & \multicolumn{1}{c|}{ {73.08}} & \multicolumn{1}{c|}{85.40} & \multicolumn{1}{c|}{75.37} & \multicolumn{1}{c|}{82.64} & \multicolumn{1}{c}{\bf{86.71}}\\
\bottomrule
\end{tabular}}
\vspace{-2mm}
\end{table}

\subsection{Ablation Study}
\textbf{Effectiveness of multi-view encoding.~}
Tab.~\ref{tab2} demonstrates the effect of various \textit{multi-view encoding} components, alongside evaluations conducted on Gym99 using a $5-$way $1-$shot setup without \textit{multi-view fusion}. Our observations reveal that individual components of \textit{multi-view encoding} play a vital role in view-specific matching, displaying distinguishable characteristics under different combinations. Notably, intra-frame context encoding, denoted by {\bf I-F}, improves instance-specific matching (\emph{i.e.,~}{\bf I-M}) by $7.65\%$, intra-video context encoding, denoted by {\bf I-V}, improves category-specific matching (\emph{i.e.,~}{\bf C-M}) by $6.59\%$, and intra-episode context encoding, denoted by {\bf I-E}, improves task-specific matching (\emph{i.e.,~}{\bf T-M}) by $30.67\%$. Furthermore, the inputs of {\bf I-V} and {\bf I-E} are first augmented by {\bf I-F} to enhance feature discriminability in M$^3$Net, as displayed in Fig.~\ref{fig2:overview}. The proposed {\bf I-F} further improves " {\bf I-V+C-M} " and " {\bf I-E+T-M} " by $0.88\%$ and $2.82\%$, respectively, indicating the significance of enriching spatial context within each frame before modeling the temporal and task relationships.

\noindent
\textbf{Influence of multi-view matching.~}
Our proposed \textit{multi-view matching} aims to precisely identify the most informative frame correspondences that highlight the similarity between the video pairs. We perform a comprehensive analysis of different \textit{multi-view matching} variants by varying their constituent components, as outlined in Tab.~\ref{tab4}. We observe that the different combinations have distinct properties, \emph{e.g.,~} {\bf C-M} is more effective for temporal matching than {\bf I-M} and {\bf T-M}, thanks to the cross-attention mechanisms that operate over the query and support features. When the enriched embeddings are sufficiently discriminative, the performance of {\bf I-M} and {\bf T-M} is similar. However, we observe a slight degradation in the overall performance in the last column of Tab.~\ref{tab4} when we swap the positions of {\bf C-M} and {\bf T-M}, even though both matching functions are \textit{non-temporal}, indicating the efficacy of the tailor-made matching designs for different higher-order embeddings obtained from the \textit{multi-view} encoding.

\begin{table}[t!]
\caption{\small Comparison between various combinations in multi-view fusion for 5-way 1-shot action recognition on Gym99.}\label{tab3}
\vspace{-1mm}
\resizebox{0.95\linewidth}{!}{
\begin{tabular}{cllllllll}
\toprule
\multicolumn{2}{c|}{\bf Model Index} & \multicolumn{1}{c}{$\mathbf{1}$} & \multicolumn{1}{c}{$\mathbf{2}$} & \multicolumn{1}{c}{$\mathbf{3
}$} & \multicolumn{1}{c}{$\mathbf{4}$} & \multicolumn{1}{c}{$\mathbf{5}$} & \multicolumn{1}{c}{$\mathbf{6}$} & \multicolumn{1}{c}{\bf $\mathbf{7}$}\\
\midrule
\multicolumn{2}{c|}{\bf Fusion Num} & \multicolumn{1}{c}{$1$} & \multicolumn{1}{c}{$1$} & \multicolumn{1}{c|}{$1$} & \multicolumn{1}{c}{$2$} & \multicolumn{1}{c}{$2$} & \multicolumn{1}{c|}{$2$} & \multicolumn{1}{c}{\bf $3$}\\
\midrule
%L1
\multicolumn{1}{c|}{\multirow{2}*{\bf Multi-view}}&\multicolumn{1}{c|}{$\mathcal{L}_1$}& \multicolumn{1}{c}{\bf \checkmark}&\multicolumn{1}{c}{-}&\multicolumn{1}{c|}{-}& \multicolumn{1}{c}{-} & \multicolumn{1}{c}{\bf \checkmark} &\multicolumn{1}{c|}{\bf \checkmark}& \multicolumn{1}{c}{\bf \checkmark}\\
%L2
\multicolumn{1}{c|}{\multirow{2}*{\bf Loss}}&\multicolumn{1}{c|}{$\mathcal{L}_2$} & \multicolumn{1}{c}{-}&\multicolumn{1}{c}{\bf \checkmark}&\multicolumn{1}{c|}{-}& \multicolumn{1}{c}{\bf \checkmark} & \multicolumn{1}{c}{-} &\multicolumn{1}{c|}{\bf \checkmark}& \multicolumn{1}{c}{\bf \checkmark}\\
%L3
\multicolumn{1}{c|}{}&\multicolumn{1}{c|}{$\mathcal{L}_3$} & \multicolumn{1}{c}{-}&\multicolumn{1}{c}{-}&\multicolumn{1}{c|}{\bf \checkmark}& \multicolumn{1}{c}{\bf \checkmark} & \multicolumn{1}{c}{\bf \checkmark} &\multicolumn{1}{c|}{-}& \multicolumn{1}{c}{\bf \checkmark}\\
\midrule
% Y1
\multicolumn{1}{c|}{\multirow{2}*{\bf Multi-view}}&\multicolumn{1}{c|}{$\mathbf{Y}_1$} &\multicolumn{1}{c}{ {50.09}}&\multicolumn{1}{c}{-}& \multicolumn{1}{c|}{-} &\multicolumn{1}{c}{-}& \multicolumn{1}{c}{ {57.93}} & \multicolumn{1}{c|}{{65.59}} & \multicolumn{1}{c}{ \bf{69.35}}\\
% Y2
\multicolumn{1}{c|}{\multirow{2}*{\bf Prediction}}&\multicolumn{1}{c|}{$\mathbf{Y}_2$} &\multicolumn{1}{c}{-}& \multicolumn{1}{c}{ {69.53}} &\multicolumn{1}{c|}{-}& \multicolumn{1}{c}{ {70.21}} &\multicolumn{1}{c}{-}& \multicolumn{1}{c|}{ {70.60}} & \multicolumn{1}{c}{ \bf{71.82}}\\
%Y3
\multicolumn{1}{c|}{}&\multicolumn{1}{c|}{$\mathbf{Y}_3$} & \multicolumn{1}{c}{-}&\multicolumn{1}{c}{-}&\multicolumn{1}{c|}{ {59.79}}& \multicolumn{1}{c}{ {68.51}} & \multicolumn{1}{c}{ {59.76}} &\multicolumn{1}{c|}{-}& \multicolumn{1}{c}{ \bf{69.89}}\\
\midrule
\multicolumn{1}{c|}{\bf Fusion}&\multicolumn{1}{c|}{$\mathbf{Y}$} & \multicolumn{1}{c}{ {50.09}} & \multicolumn{1}{c}{ {69.53}} & \multicolumn{1}{c|}{ {59.79}} & \multicolumn{1}{c}{ {71.81}} & \multicolumn{1}{c}{ {60.94}} & \multicolumn{1}{c|}{ {71.47}} & \multicolumn{1}{c}{ \bf{72.70}}\\
\bottomrule
\end{tabular}}
\vspace{-2mm}
\end{table}

\begin{figure}[t!]
	\centering
	\begin{tabular}{c}
	\includegraphics[width=\linewidth]{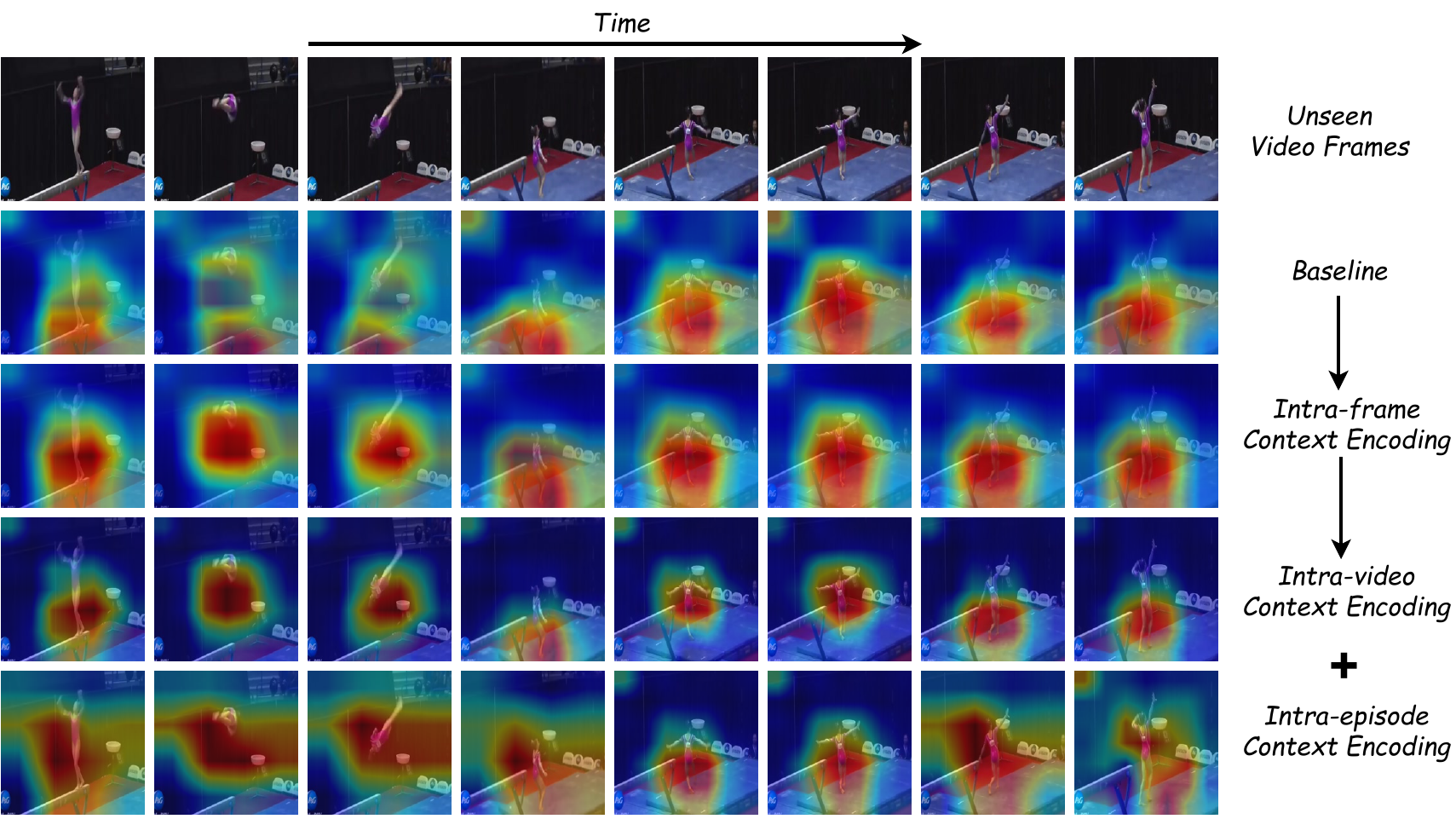} \\
	\end{tabular}
	\vspace{-2mm}
	\caption{\small The comparison of activation map visualizations for one unseen example from the Gym99 test set. 
 }%
	\label{fig5}
 \vspace{-2mm}
\end{figure}

\noindent
\textbf{Analysis of multi-view fusion.~}
Further experiments are conducted on various combinations of \textit{multi-view fusion}, and the obtained results, presented in Tab.~\ref{tab3}, revealed the following: (i) The performance of \textit{multi-view fusion} improves as the matching view diversity increases. (ii) Incorporating multiple matching functions through \textit{multi-view fusion} provides superior performance compared to individual views, thus attaining a more comprehensive decision maker. As illustrated in the last column of Tab.~\ref{tab3}, Model-$\mathbf{7}$ with multi-view prediction fusion record $72.70\%$ for $\mathbf{Y}$, surpassing the corresponding three single-view predictions (\emph{i.e.,~}$69.35\%$ for $\mathbf{Y}_1$, $71.82\%$ for $\mathbf{Y}_2$, and $69.89\%$ for $\mathbf{Y}_3$ respectively). (iii) The single-view prediction (\emph{i.e.,~}$\mathbf{Y}_1$, $\mathbf{Y}_2$, and $\mathbf{Y}_3$) of the full model with multi-view loss fusion surpasses that of the independent model with single-view loss, and the advantage thereof increases with the view diversity. Specifically, compared to Model-$\mathbf{1}/\mathbf{2}/\mathbf{3}$, Model-$\mathbf{7}$ exhibits an improvement in the performance of all single-view predictions from $50.09\%$, $69.53\%$, and $59.79\%$ to $69.35\%$, $71.82\%$, and $69.89\%$, respectively. These results demonstrate that the proposed \textit{multi-view fusion} guides the full model to attain more discriminative embeddings and thereby surpasses corresponding independent models.

% \vspace{-1mm}
\subsection{Visualization results}
\begin{figure}[t!]
	\centering
	\begin{tabular}{c}
	\includegraphics[width=\linewidth]{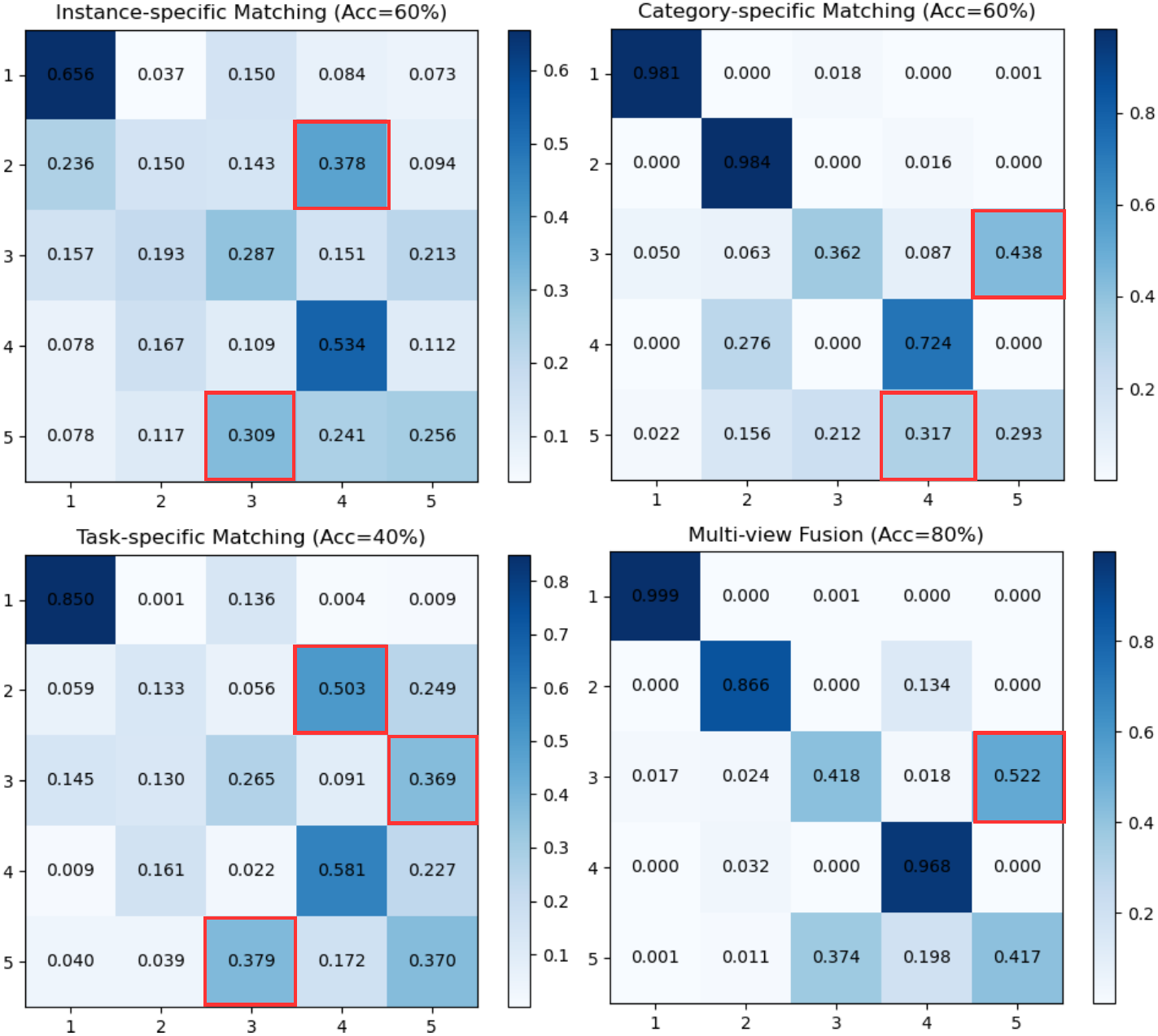} \\
	\end{tabular}
	\vspace{-2mm}
	\caption{\small The prediction distribution shows how query videos (columns) match to support videos (rows) in robust multi-view fusion. The \textcolor{red}{red} box indicates the incorrect match.}%
	\label{fig6}
 \vspace{-2mm}
\end{figure}

Fig.~\ref{fig5} presents the progressive integration of our contributions in \textit{multi-view encoding} from top to bottom, highlighting the distinctiveness of the learned context-enriched features. 
The integration of IFCE (third row) enhances the spatial representation of features and induces attention toward relevant objects in a single video frame. 
After this integration, the integration of IVCE (fourth row) further strengthens the temporal relation of features, allowing our M$^3$Net to focus on the action subject while reducing attention to extraneous objects. 
Meanwhile, the integration of IECE (fifth row) amplifies and emphasizes task-specific regions with higher saliency, which may be more relevant for query predictions, while still retaining relatively complete original information.
Additionally, Fig.~\ref{fig6} qualitatively shows the robustness of the proposed \textit{multi-view fusion}. We present the prediction distribution emanating from various matching functions for a novel $5-$way $1-$shot task and visually demonstrate the effectiveness of the proposed \textit{multi-view fusion}. 
% Obviously, \textit{multi-view fusion} leads to a more precise prediction. 

\section{Conclusion}
In this paper, we have proposed M$^3$Net, a matching-based framework for FS-FG action recognition. The proposed framework leverages a multi-view encoding procedure, capturing rich contextual information across intra-frame, intra-video, and intra-episode views to generate customized representations. Moreover, M$^3$Net adopts a multi-task collaborative learning paradigm, integrating instance-specific, category-specific, and task-specific matching functions to model complex spatial and temporal relations, thereby achieving remarkable performance improvements over current state-of-the-art methods on three challenging fine-grained action recognition benchmarks. Our findings provide valuable insight for future research to exploit the \textit{structural invariance} of multiple views to capture subtle spatial semantics and complex temporal dynamics for fine-grained video understanding, especially under the condition of data scarcity. This has broad practical implications for multimedia domains such as sports analytics and video surveillance.